\documentclass[conference]{IEEEtran}
\IEEEoverridecommandlockouts
\usepackage{cite}
\usepackage{amsmath,amssymb,amsfonts}
\usepackage{algorithmic}
\usepackage{graphicx}
\usepackage{textcomp}
\usepackage{xcolor}

\def\BibTeX{{\rm B\kern-.05em{\sc i\kern-.025em b}\kern-.08em
    T\kern-.1667em\lower.7ex\hbox{E}\kern-.125emX}}
\begin{document}

\title{Solving the Reaction-Diffusion equation based on analytical methods and deep learning algorithm; the Case study of sulfate attack to concrete\\
}

\author{
\IEEEauthorblockN{Amin Karimi Monsefi}
\IEEEauthorblockA{\textit{Computer Science and Engineering} \\
\textit{Shahid Beheshti University}\\
Tehran, Iran \\
a\_karimimonsefi@sbu.ac.ir}
\and
\IEEEauthorblockN{Rana Bakhtiyarzade}
\IEEEauthorblockA{\textit{Metallurgy and Materials Engineering} \\
\textit{Iran University of Science and Technology}\\
Tehran, Iran \\
rana\_bakhtiyarzade@metaleng.iust.ac.ir}

}

\maketitle

\begin{abstract}
The reaction-diffusion equation is one of the cornerstones equations in applied science and engineering. In the present study, a deep neural network has been trained in order to predict the solution of the equation with different coefficients using the numerical solution of this equation and the utility of deep learning. Analytical solution of the Reaction-Diffusion equation also has been conducted by taking advantage of Danckwert’s method. The accuracy of deep learning results was compared with the analytical solutions. In order to decrease the learning time and to find out similar equations’ solutions, such as pure diffusion and pure reaction, dimensional analysis technique has been performed. It was demonstrated that deep learning can accurately estimate the Partial Differential Equation’s solution in the case of the reaction-diffusion equation with a constant coefficient.
\end{abstract}


\section{Introduction}
Partial Differential Equations (PDEs) have always played a significant role in mathematical modeling and simulation, which are widely used in physics, engineering and economics \cite{morton2005numerical,zakeri2019weakly}. A wide range of physical phenomena are modelled by second-order PDEs with constant coefficients, and representing precise solutions for these kinds of equations is an important part of applied mathematics. Although analytical and numerical methods have been widely developed for solving PDEs, each one has its own advantages and drawbacks. While analytical methods lead to precise solutions, they are not applicable for most of PDEs, and by any changes in boundary and initial conditions type, they become useless. On the other hand, numerical methods enable scientists to solve complex PDE problems, but these methods are mesh dependent and have high computational costs, and also their solutions are not reliable before validation with analytical solutions or experimental results. In order to reach a fast and accurate technique, which utilizes the advantages of both numerical and analytical methods, a mesh-free tool is needed. Consequently, an intelligent method which enable us to comprehend the analytical solutions and generalize them for complex geometries is the best choice.

Deep learning is a part of artificial intelligence which is quite useful and provide several advances in finding the solution of high dimensional data problems. In recent years several pieces of research have demonstrated significant success in artificial intelligence \cite{goodfellow2016deep,lecun2015deep,krizhevsky2012imagenet,hinton2012deep,silver2016mastering,bai2009novel,liu2018variational,sharma2018weakly,sirignano2018dgm,zhou2017brief}. Although the idea of using the neural network is an old idea, in recent years researches show that a wide variety of excellent modeling of complicated data sets were established using the mentioned method \cite{han2018solving}. Deep learning methods as a representation learning method use several levels, and each level uses non-linear schemes to representation at a higher level. By using enough transformations, complicated functions also be learned. It is noticeable that in deep learning method all of the discussed levels are designed by computer itself \cite{lecun2015deep,monsefi2019performing}.

In this work, a deep neural network algorithm has been trained by using the numerical solution of a specific PDE in finite intervals. The purpose of this network is to learn the behaviour of the equation for predicting the values of the solution for whole domain. By taking the advantage of the analytical solution, error and the accuracy of deep learning result was calculated.

Reaction-Diffusion equation is one of the most famous PDEs in engineering problems which is used as a case study in this paper. This equation is widely used for prediction of sulfate concentration in concrete during the sulfate attack process \cite{zuo2012numerical,cefis2017chemo,sun2013new}. The one-dimensional Reaction-Diffusion equation with Dirichlet boundary condition has been solved analytically on symmetric line. In order to solve the analytical solution, firstly the pure diffusion equation solved, then based on danckwert’s transformation the reaction-diffusion solution would be extracted from the pure Diffusion equation solution \cite{crank1979mathematics}. The Diffusion and the Rate of Reaction coefficients for the case of sulfate attack in concrete have assumed based on the Zuo et al work \cite{zuo2012numerical}. The numerical solution of the discussed equation has also extracted by the second-order algorithm of finite difference method. A deep neural network with several main hidden layers designed to learn the behavior of the equation from the numerical results, and all the influential parameters on the solution of the equation classified into different categories, and considered as a separate input parameter for the deep learning algorithm. Both the solution and parameters were used as feed for the first layer of the algorithm and trained the neural network to predict the solution of the equation with logical possible parameters for the problem. In order to have the better judgment about the solution, a practical case study of sulfate attack which was modeled by Zuo et al \cite{zuo2012numerical} was used, and all the coefficients and boundary and the initial conditions were set similar to their research. The results which have calculated by the deep learning algorithm has compared to analytical results, and by using dimensional analysis methods, the solutions of pure reaction and pure diffusion was also extracted from deep learning results. The deep learning based solution is quite similar to the analytical methods with low error and high accuracy.

\section{Methodology}

\subsection{Governing Equations}
The governing equation for all the transportation phenomena including sulfate attack can be expressed as the following
equation \cite{versteeg2007introduction}:

\begin{equation}
 \frac{\partial  \rho  \phi }{\partial t} +    \nabla  ( \rho \overrightarrow{ \upsilon }\phi  ) =   \nabla (  \Gamma _{ \phi }    \nabla   \phi  ) +  S_{ \phi }    
\label{eq_1}
\end{equation}

where  $\phi$  is the generalized scalar, $\Gamma _{ \phi } $ is global diffusion coefficient, $S _{ \phi }$ is the source term and,  $\overrightarrow{ \upsilon }$  and $ \rho $ are the velocity and
density of the fluid respectively.

It is assumed that there is no fluid in the case of sulfate attack, as a result, the convection term of the governing Eq.\ref{eq_1} is omitted from the main equation. so it will be simplified as follows:

\begin{equation}
 \frac{\partial  \phi }{\partial t}  =  \nabla (  \Gamma _{ \phi } \nabla  \phi  ) +  S_{ \phi }   
\label{eq_2}
\end{equation}

Zuo \cite{zuo2012numerical} has reformed the governing equation using sulfate attack phenomena parameters which are represented in the following table:

\begin{table}[h]
 \caption{Coefficients of equation. \label{table_1}}
 \begin{center}
 \begin{tabular}{c c c c} 
 \hline
 Equation & $S_{ \phi }$ & $ \phi $ & $\Gamma _{ \phi }$ \\ [0.5ex] 
 \hline
 reaction\_diffusion & $  \frac{\partial  C_{d} }{\partial t} $ & $C$ & $D_{e}$ \\  [1ex] 
 \hline
\end{tabular}
\end{center}
\end{table}

Eventually the transport equation considering boundary and initial conditions and with the assumption of one-dimensional
diffusion, the corresponding equation will be reformed to the form of Eq.\ref{eq_3}:

\begin{equation}
  \ x =\begin{cases}  \frac{\partial C}{\partial t} =  \frac{\partial  (D_{e}(x,t)  \frac{\partial C}{\partial x} ) }{\partial x} +  \frac{\partial C_{d}}{\partial t}     \\ C(X,0) = 0   & x  \in [0,L]  \\ C(0,t) = C_{0} & C(L,t) =  C_{0}  \end{cases}   
\label{eq_3}
\end{equation}

Where $C$ is the concentration of the sulfate $ions(mol/m3)$, $x$ is the location of the section$(m)$, t is the time of diffusion$(s)$, $D_{e}$ is the Ionic diffusion coefficient of sulfate in concrete $(m2/s)$, $C_{d}$ is the dissipation concentration which is caused by the chemical reactions of sulfate $ions(mol/m3)$, $C_{0}$ is the boundary concentration of sulfate $ions(mol/m3)$, L is the thickness of the concrete member, and $[0,L]$ interval represents the concrete cross section \cite{zuo2012numerical}.

\subsubsection{Chemical Reactions Rate of Sulfate Dissipation}

Reaction-diffusion phenomena is a composition of chemical reactions and ionic diffusivity at the same time which in order
to solve the equation, each one of these terms must be evaluate and calculate individually. Reaction term of the Eq.\ref{eq_2} can be expressed as:

\begin{equation}
 \frac{\partial C_{}}{\partial t}  = -k_{v} . C_{ca^{2+} }. C 
\label{eq_4}
\end{equation}

Where $K_{v}$ is the rate of chemical reaction in Eq.\ref{eq_4}; $C_{ca^{2+} }$ is the calcium ion concentration in pore solution, and the value of $k_{v} . C_{ca^{2+} }$ is assumed to be the rate of reaction $(k)$. This can be assumed as a linear function of temperature and is taken as $25 mol/m3$ a $273 K$ and $10 mol/m3$ at $373 K$ \cite{perry1950chemical}.

\subsubsection{Diffusion Coefficie}
There are $3$ main methods in calculation of diffusion coefficient, considering diffusion coefficient as a:

\begin{itemize}
	\item Constant value, as Zuo represented his numerical method.
	\item A time variable value which is explained in Sun’s paper \cite{sun2012time}.
	\item A time-depth variable value which is the most accurate method \cite{zuquan2007interaction}.
\end{itemize}

Since there is not much difference between the accuracy of these methods and the first method has an acceptable estimation
of sulfate ions diffusion, the so first method is utilized to find an analytical solution. By the above assumption the general equation will reform as follows:

\begin{equation}
  \frac{\partial C}{\partial t} = D_{e} \frac{\partial^2C}{\partial x^2} - kC  
\label{eq_5}
\end{equation}

Where $C$ is a concentration of the substance, $kC$ is the rate of removal of diffusing substance and $k$ is the rate of reaction which is a constant value.

\subsection{Analytical Solution}

In order to solve Eq.\ref{eq_5} Danckwert’s(1951) has tried a specific method which by the use of a simple transformation reform the reaction-diffusion equation to a case that there would be diffusion without reaction. He has presented his method based on to different types of boundary conditions which are \textit{Dirichlet} and \textit{Neumann}. In the case of sulfate attack, the \textit{Neumann} boundary condition has utilized. \textit{Neumann} boundary conditions are represented as follows:

\begin{center}
  $C = 0$,  $t = 0$,  at all points in the medium
\end{center}
and also
\begin{center}
  $C = C_{0}$,  $t > 0$,  at all points on the surface
\end{center}
In the next step by the assumption of no reaction, $C_{1}$ will be the solution of the general equation.

\begin{equation}
   \frac{\partial C_{1}}{\partial t} = D \frac{\partial^2C_{1}}{\partial x^2}  
\label{eq_6}
\end{equation}
The answer of the following equation will give the actual value of $C$:

\begin{equation}
   C =  \int_0^t C_{1} e^{-k  t' } d t' + C_{1} e ^ {-kt}     
\label{eq_7}
\end{equation}
Finally with solving the Eq.\ref{eq_6} and pasting in Eq.\ref{eq_8} and by doing the calculations with the boundary conditions of Eq.\ref{eq_3} the final solution will be expressed as follows:

\begin{equation}
  C(x,t)  = -  \frac{4C_{0}}{ \pi }  \sum_{n=0}^{ \infty } (a_{n} cos( \omega _{n}x)(k \Psi _{n}( p - 1) + p) k) + C_{0}
\label{eq_8}
\end{equation}
Where $a_{n}$, $ \omega_{n}$, $ \Psi_{n}$ and $p$ are represented as follows:

\begin{center}
  $a_{n} =  \frac{(-1)^{n}}{2n + 1} $,  $ \omega _{n} =  \frac{(2n+1) \pi }{2L} $,  $ \Psi _{n} =  \frac{4L^2}{(-D_{e}(2n+1)^2 \pi ^2) - 4kL^2} $, $p = exp( \frac{t}{ \Psi _{n}} )$
\end{center}

\subsection{Deep lLearning Method}
In this section we introduce our presented deep neural network techniques to find a solution for the reaction-diffusion equation,
with the use of produced data, we attempted to teach the neural network.

\subsubsection{Production data}
To solve the intended reaction-diffusion equation with the use of deep neural networks it is essential to teach the neural
network with specific parameters, for the computers to understand the discussed PDE it is needed to extract the critical features of the equation. Since the partial differential equations are defined on the continuous space-time, their domain can be varied from zero to infinity while the computers understand discrete mathematics and limited domains so it is not possible for the computers to understand the issue, Hence it is needed to discretize the period of the features for the computer to understand the function. The features and their period are introduced in the following table:

\begin{table}[h]
 \caption{Discretization intervals and the dimension of T is in the scale of per year. \label{table_2}}
 \begin{center}
 \begin{tabular}{c c c c c c} 
 \hline
 $C$ & $L$ & $ X $ & $T$ & $k$& $D_{e}$ \\ [0.5ex] 
 \hline
 0 - 200 &  0 - 0.05 & -L - L & 0 - 7 & $ \frac{1}{10}- \frac{1}{10^{10}}$& $\frac{1}{10}- \frac{1}{10^{13}}$ \\  [1ex] 
 \hline
\end{tabular}
\end{center}
\end{table}

where $C$ is concentration, $L$ is half of the domain, $X$ is the variable position on the domain, $T$ is the time (year), $k$ is reaction rate and $D_{e}$ is the effective diffusion coefficient.

By the use of \textit{gaussian random} data generation method, 3 million data has produced. After the data production, data were
categorized into categories which are called Batch, and 1000 Batches has been produced which each one contains 3000 data. The main idea of categorizing process is to increase the learning model rate. Specific conditions are considered for each parameter to optimize the data distribution so the range of each parameter is discretized into different parts, the data in each part are chosen randomly with specific steps. The selected steps differ for each part of the input data.

Like the table above, the rest of the parameters based on the modality of each parameter they will be classified and then the
data generation process will be done. The input data is made of different parameters which each one of them covers a specific period so, in order to increase the learning rate, the data will be normalized with the following methods:

\begin{equation}
   \mu =  \frac{1}{m}  \sum_{i=0}^{n} X_{i},   \sigma^{2} =  \frac{1}{m} \sum_{i=0}^{n} X_{i}^{2},  x' = x -  \mu,  x'' =  \frac{x'}{ \sigma ^2} 
\label{eq_9}
\end{equation}

The data generated as $x”$ from Eq.\ref{eq_9} will be considered as the input data.

\subsubsection{Learning model method}

A 3-layer network is considered for the learning model in which the layers of the network are introduced as follows:
\begin{itemize}
	\item Input layer.
	\item Hidden layer.
	\item Output layer.
\end{itemize}
The input layer receives the data as a matrix and delivers them to hidden layers. Hidden layers after a process send them to the output layer.
\begin{figure}[t]
\caption{Deep neural network diagram}
\label{image_1}
\includegraphics[width=8cm]{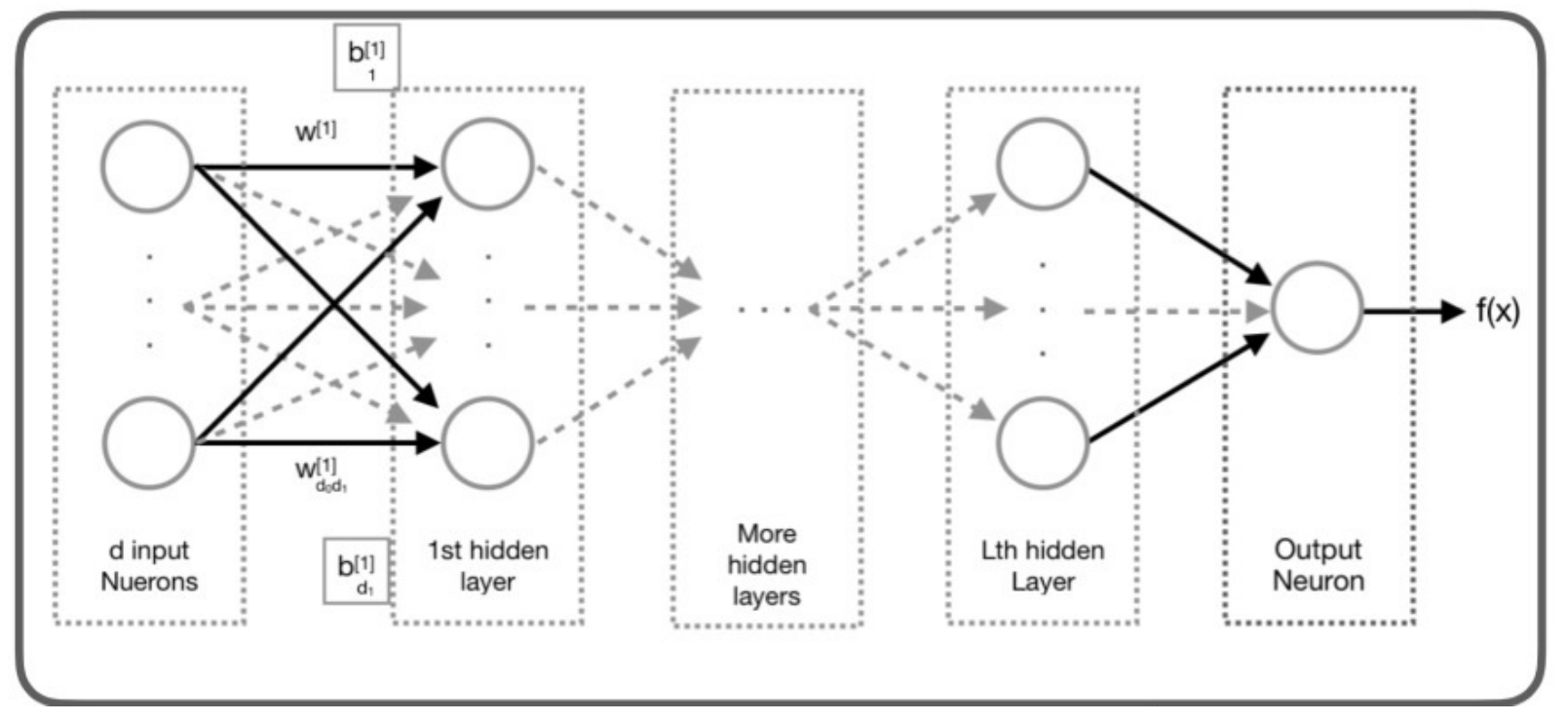}
\centering
\end{figure}

The figure \ref{image_1} is a general schematic of trained neural network. Where in this network $w^{[l]}$ is the weight matrix which relates $l$ and $l - 1$ bit-layers, $w^{[l]}_{ij}$ is the amount of matrix weight for neuron $i$ in layer $l$. $j$ is in the layer $ - 1$ and the vector $b^[l]$ is equal to bios amount for layer $l$. The equations above demonstrate the updating method of the discussed amounts.
\begin{equation}
  Z^{[l]} = W^{[l]} * A^{[l-1]} + B^ {[l]}
\label{eq_10}
\end{equation}
\begin{equation}
  A^{[l]} = g^{[l]}(Z^{[l]})
\label{eq_11}
\end{equation}
In this kind of situations $A^{[l]}$ is a matrix which is the amount of the previous layer. The function $g^{[l]}$ is the utilized activation function in the $l’$th layer the activation function which used in the hidden layer is function of LeakyReLU and Sigmoid which are explained as follows:
\begin{equation}
  Sigmoid(x)  =  \frac{1}{1 + e^{-x}} 
\label{eq_12}
\end{equation}
\begin{equation}
 LeakyReLU = \begin{cases}x & x  \geq  0\\ x * 0.001  & x < 0\end{cases} 
\label{eq_13}
\end{equation}
First, the values of $W$ will be randomly chosen between $0$ to $1$ and then the values of $B$ will be considered less than zero. The main purpose of learning model method is to decrease the error function.
\begin{equation}
   \min_{W,B} J(W,B) 
\label{eq_14}
\end{equation}
\begin{equation}
 J(W,B) =  \frac{1}{m} ||Y' - Y||^{2}_{2}  
\label{eq_15}
\end{equation}
Where $Y'$ and $Y$ are the estimated value and the exact value which is extracted from the analytical solution. To prevent the \textit{Overfitting} in this part, the error function was changed as follows:

\begin{equation}
   J(W,B) =  \frac{1}{m} ||Y' - Y||^{2}_{2}+  \frac{ \lambda }{2m} ||W||^{2}_{2} 
\label{eq_16}
\end{equation}

\begin{equation}
 J(W,B) =  \frac{1}{m}  \sum (Y' - Y)^{2} +  \frac{ \lambda }{2m} \sum_{j=1}^{n_{x}} W_{j}^{2}    
\label{eq_17}
\end{equation}

\begin{equation}
 J(W,B) =  \frac{1}{m} (Y'- Y)^{T}(Y' - Y) +  \frac{ \lambda }{2m} W^{T}W    
\label{eq_18}
\end{equation}

To increase the learning rate, the Adam’s optimization method was utilized.
\subsubsection{Tuning parameters}
One of the most critical and inevitable parts of the learning algorithm is tuning parameters. In order to achieve a better learning model, it is essential to have an acceptable estimation in the following parameters: 1-$\alpha$ (learning rate) 2-the number of hidden layers 3-the number of neurons in the hidden layer.

In the next step, it’s needed to divide the generated data into $3$ categories: \textbf{test data}, \textbf{validation data}, \textbf{training data}. $90$
percent of the $1000$ batches were allocated to the numerical solution and considered for the train, and also $5$ percent of generated data was used for production validation results by the numerical method in order to fix deep learning parameters. The $5$ percents of remaining data were used by the analytical solution to produce test results.

\section{Results}
In this section, the accuracy of the deep learning solution has analyzed in comparison to the analytical solution which by changing the critical features of the corresponding equation this could be examined. there are $2$ main strategy to examine and evaluate the solutions in the deep learning method:

\subsection{Threshold Concept}

Firstly the precision of deep learning method and it’s error with the analytical solutions of the given PDE should be evaluated then the errors must be checked in different learning data, in order to this $1000$ batches has been generated which each one contains $3000$ data. In each level of examination, the $X$ number of batches should be selected out of the $1000$ given batches. The experiment under the following conditions has been taken: $90\%$  of the given data was considered as training data, $5\%$ as validation data and $5\%$ as test data.

The \textit{Mean Square Error} index has been utilized to calculate the $3$ following errors: training error, validation error, and test error. The \textit{MSE} index is represented in Eq.\ref{eq_19}:

\begin{equation}
 MSE =  \frac{ \sum_{i=0}^{n} (y' - y)^{2} }{n}  
\label{eq_19}
\end{equation}

The \textit{Threshold concept} has utilized in order to compare the exact amount of the given PDE with the results form deep learning method. Whereas $y'$ is the deep learning calculated quantity and $y$ represents the amount of PDE.

\begin{equation}
 |y - y'| <   \theta  
\label{eq_20}
\end{equation}
If the threshold quantity was more than left-hand side of Eq.\ref{eq_20}, then both values will be assumed as equal.

\begin{table}[h]
 \caption{Batch number sensitivity analyze Batch. \label{table_3}}
 \begin{center}
 \begin{tabular}{c c c c c c} 
 \hline
 $\#Batch$ & $Training$ & $Validation$ & $Test$ & $Thr(2) $& $Thr(1)$\\ [0.5ex] 
 \hline
 $100$ & $0.8954$ & $6.1547$ & $9.1574$ & $ 71.52\% $& $ 65.14\% $ \\  [1ex] 
  $300$ & $0.8521$ & $5.9856$ & $7.1259$ & $ 74.32\% $& $ 70.78\% $ \\  [1ex] 
  $500$ & $0.7485$ & $5.0198$ & $5.1245$ & $ 78.91\% $& $ 73.15\% $ \\  [1ex] 
  $700$ & $0.6574$ & $3.1497$ & $4.1547$ & $ 85.47\% $& $ 80.19\% $\\  [1ex] 
  $1000$ & $0.5782$ & $1.8643$ & $3.0214$ & $ 91.71\% $& $ 89.18\% $ \\  [1ex] 
 \hline
\end{tabular}
\end{center}
\end{table}

Looking at table.\ref{table_3} in more details, by increasing the number of batches training and test errors have been decreased. it is also noticeable that the validation data fall by increasing the number of batches. during the learning process the main aim is to decrease the validation data error which have been satisfied properly.moreover, increment of the batches directly influence the accuracy of the deep learning process.

\subsection{Changing Parameter Analyze}
In this section, the dependency of the deep neural network to the value of the equation’s coefficients have been analyzed. In
order to conduct this purpose, all of PDE’s coefficients were considered as constant values except one of them, and by changing that coefficient, the accuracy of the deep learning solution has been compared to the analytical solution with using three different thresholds.

The variable parameters in this part are $k$ and $D_{e}$
which play a central role in the behavior of the equation. Table.\ref{table_4} 
demonstrate the accuracy of deep learning solution by changing the $k$ and $D_{e}$ values, where $C_{0}$ and $L$ are $75.5$ and $0.05$ respectively.

\begin{table}[h]
 \caption{Accuracy analyze of deep learning solution based on changing coefficie. \label{table_4}}
 \begin{center}
 \begin{tabular}{c c c c} 
 \hline
 $k$ & $Thr(2)$ & $Thr(1)$ & $Thr(0.5)$ \\ [0.5ex] 
 \hline
 $2.125 * 10^{-2}$ & $ 87.56\%$ & $82.39\%$ & $79.45\%$ \\  [1ex] 
  $2.125 * 10^{-5}$ & $ 87.84\%$ & $83.05\%$ & $80.25\%$ \\  [1ex] 
  $2.125 * 10^{-7}$ & $ 88.69\%$ & $84.58\%$ & $82.12\%$ \\  [1ex] 
  $2.125 * 10^{-10}$ & $ 80.32\%$ & $78.73\%$ & $77.45\%$ \\  [1ex] 
  $2.125 * 10^{-13}$ & $ 64.78\%$ & $63.58\%$ & $61.03\%$ \\  [1ex] 
 \hline
\end{tabular}
\small $D_{e} = 2.6 \times 10^{-9}$
\end{center}

 \begin{center}
 \begin{tabular}{c c c c} 
 \hline
 $D_{e}$ & $Thr(2)$ & $Thr(1)$ & $Thr(0.5)$ \\ [0.5ex]
 \hline
 $2.6 * 10^{-5}$ & $ 73.38\% $ & $66.39\%$ & $61.45\%$ \\  [1ex] 
  $2.6 * 10^{-7}$ & $ 84.29\% $ & $82.48\%$ & $76.21\%$ \\  [1ex] 
  $2.6 * 10^{-10}$ & $ 89.29\% $ & $87.78\%$ & $82.27\%$ \\  [1ex] 
  $2.6 * 10^{-12}$ & $ 88.57\% $ & $86.54\%$ & $82.12\%$ \\  [1ex] 
  $2.6 * 10^{-15}$ & $ 81.33\% $ & $75.91\%$ & $70.67\%$ \\  [1ex] 
 \hline
\end{tabular}
\small $k = 2.125 \times 10^{-7}$
\end{center}
\end{table}

According to the table.\ref{table_4}, it is understood that the accuracy of the solution strongly depends on the correlation of \textit{Rate of reaction} and \textit{Diffusion coefficient}. In the case of constant \textit{Diffusion coefficient} where $k$ value is less than the particular amount the accuracy of the solution drop considerably. It is also similar for the case of the constant \textit{Rate of reaction} where if $D_{e}$ value chose larger than specific value, the solution will not be reliable. When it comes to physics of the phenomena, reduction of the accuracy of the deep learning solution become reasonable. This is because our network was trained for Reaction-Diffusion equation but in these intervals, the equation is changed to the pure reaction and pure Diffusion. To tackle this issue using Dimensional Analysis methods would be useful.

\subsection{Dimensional Analyze}

In order to conduct the pure reaction and Diffusion equation it was needed to change the Eq.\eqref{eq_2} to a dimensionless form which
is expressed as follows:

\begin{equation}
  \frac{\partial C^{*}}{\partial t^{*}} =  \frac{D_{c}t_{c}}{L^{2}} *  \frac{\partial^2C^{2}}{\partial x^{*2}} - kt_{c}C^{*}     
\label{eq_21}
\end{equation}
where dimensionless parameters are defined as follows :

\begin{center}
    $x^{*} = \frac{x}{L}$, $t^{*} = \frac{t}{t_{c}}$, $C^{*} = \frac{C}{C_{0}}$
\end{center}

The reason behind using the dimensional analyze is that we want to find out the solution of the pure Reaction and Diffusion
equation with a deep learning algorithm which is designed for Reaction-Diffusion Equation. For this purpose, a famous dimensionless number was used for finding suitable coefficients and then, the solution compared to the analytical solution of the pure Reaction equation as a case study.

\subsubsection{Damkohler number}
In reaction-diffusion phenomena there is an important dimensionless parameter which is called \textit{Damkohler number} that is
defined as follows:

\begin{equation}
  D_{a}  = \frac{Rate of reaction}{Diffusion rate}   
\label{eq_22}
\end{equation}
For Eq.\eqref{eq_21} which is the dimensionless form of the reaction-diffusion equation, Damkohler number is defined as:
\begin{equation}
  D_{a}  = \frac{kL^{2}}{D_{c}}   
\label{eq_23}
\end{equation}

This number represents the states of reaction-diffusion in different states where $D_{a}  \cong 1$, $D_{a}   \gg  1$ , and $D_{a} \ll 1$ mean the physics of Reaction-Diffusion, pure Reaction, and pure Diffusion respectively.

In the following table, by using the Damkohler number variations, the results of deep learning method with analytical solutions of pure reaction equation has been compared and the errors have calculated:

\begin{table}[h]
 \caption{Damkohler variation for reaction equation. \label{table_6}}
 \begin{center}
 \begin{tabular}{c c c c c } 
 \hline
 $D_{e}$ & $MSE$ & $ Thr(0.5) $ & $Thr(1)$ & $Thr(2)$ \\ [0.5ex] 
 \hline
 $2* 10 ^{-14}$ & $0.256$ & $94.23\%$ & $96.95\%$ & $ 98.45\%$\\  [1ex] 
 
 $2* 10 ^{-13}$ & $1.731$ & $91.14\%$ & $95.92\%$ & $ 98.03\%$\\  [1ex] 
 
 $2* 10 ^{-12}$ & $5.065$ & $90.86\%$ & $95.53\%$ & $ 97.75\%$\\  [1ex] 

 $2* 10 ^{-11}$ & $9.077$ & $90.40\%$ & $94.93\%$ & $ 96.21\% $\\  [1ex] 
 
 $2* 10 ^{-10}$ & $11.623$ & $85.32\%$ & $86.93\%$ & $ 88.36\%$\\  [1ex] 
 \hline
\end{tabular}
\bigskip
$C_{0} = 75.5(mol/m3)$ ,$L = 0.05(m)$ ,$k = 2 \times 10^{-4}(1/s)$
\end{center}
\end{table}

It is needed to note that based on the Damkohler concept, the values for Damkohler must be chosen carefully between specific
intervals. In the table.\ref{table_6} it is clear that the accuracy of deep learning solution dramatically decreases when the Damkohler number gets smaller from a specific value.

\section{Conclusion}
An analytical solution for the Reaction-Diffusion equation was conducted with assumption for the case of the sulfate attack to the concrete,and by using the numerical solution of the equation which is simply available, a deep neural network was trained to predict the behavior of the equation. The pattern recognition feature of the deep learning successfully predict the solution. It is also found that by taking the advantage of dimensional analyze it is possible to estimate the pure Reaction and Diffusion equations with good accuracy.

\bibliographystyle{./bibliography/Solving}
\bibliography{./bibliography/Solving}

\end{document}